# Advancement of MSA-technique for stiffness modeling of serial and parallel robotic manipulators

Alexandr Klimchik[1], Damien Chablat[2,3] and Anatol Pashkevich[3,4]

[1] Innopolis University, Universitetskaya 1, 420500 Innopolis, Russia
[2] Centre National de la Recherche Scientifique (CNRS), France
[3] Le Laboratoire des Sciences du Numérique de Nantes (LS2N), France
[4] IMT Atlantique, 4 rue Alfred-Kastler, Nantes 44307, France

**Abstract.** The paper presents advancement of the matrix structural analysis technique (MSA) for stiffness modeling of robotic manipulators. In contrast to the classical MSA, it can be applied to both parallel and serial manipulators composed of flexible and rigid links connected by rigid, passive or elastic joints with multiple external loadings. The manipulator stiffness model is presented as a set of basic equations describing the link elasticities that are supplemented by a set of constraints describing connections between links. These equations are aggregated straightforwardly in a common linear system without traditional merging of the matrix rows and columns, which allows avoiding conventional manual transformations at the expense of numerical inversion of the sparse matrix of higher dimension.

## 1. Introduction

In many modern robotic applications, manipulators are subject to essential external loadings that affect the positioning accuracy and provoke non-negligible compliance errors [1]. For this reason, manipulator stiffness analysis becomes one of the most important issues in the design of robot mechanics and control algorithms. It allows the designer to achieve required balance between the dynamics and accuracy. However, to make the stiffness analysis efficient, it should rely on a simple and computationally reasonable method that is able to deal with flexible links and rigid connections, passive and elastic joints that are common for serial and parallel robots.

At present, there exist three main techniques in this area [2]: Finite Element Analysis (FEA), Matrix Structural Analysis (MSA) and Virtual Joint Modeling (VJM). The most accurate of them is the FEA [3], which allows modeling links and joints with their true dimension and shape. However, it is usually applied at the final design stage because of the high computational expenses. The MSA is considered as a compromise technique, which incorporates the main ideas of the FEA, but operates with larger elements. This obviously leads to the reduction of the computational expenses, but it requires some non-trivial actions for including of passive and elastic joints in the related mathematical model.

Some reviews of existing works on manipulator stiffness analysis can be found in [4-8], where the authors addressed different aspects of the FEA, MSA and VJM techniques. Among recent contributions devoted to the MSA, it is worth mentioning the work of Cammarata [9], who merged together flexibilities of the link and adjacent joints but did not avoid manual work of lines/columns merging in the global stiffness matrix. There are also several works that deal with the MSA application to particular parallel and serial manipulators [10, 11].

The most essential contribution to the robot-oriented modification of the MSA was done by Deblase et. all [12]. The authors proposed a general technique to take into account passive joints and rigid connections via the additional constraints. Nevertheless, some manual procedures of merging matrix components were not avoided, as well as preloadings and elastic connections were not treated. For this reason, this paper focuses on the advancement of the MSA technique for robotic applications allowing to reduce related manual operations.

## 2. MSA models of manipulator links

**Modeling of the flexible link.** If the link flexibility is non-negligible, the 2-node "free-free" stiffness model should be used, which can be obtained either from the approximation of the link by a beam or using the CAD-based technique [13]. In both cases, the link is described by the linear matrix equation

$$\begin{bmatrix} \mathbf{W}_i \\ \mathbf{W}_j \end{bmatrix} = \begin{bmatrix} \mathbf{K}_{11}^{(ij)} & \mathbf{K}_{12}^{(ij)} \\ \mathbf{K}_{21}^{(ij)} & \mathbf{K}_{22}^{(ij)} \end{bmatrix}_{12 \times 12} \cdot \begin{bmatrix} \Delta \mathbf{t}_i \\ \Delta \mathbf{t}_j \end{bmatrix} \qquad (1)$$

where $\Delta \mathbf{t}_i$, $\Delta \mathbf{t}_j$ are the deflections at the link ends, $\mathbf{W}_i$, $\mathbf{W}_j$ are the link end wrenches, $i$ and $j$ are the node indices corresponding to the link ends, and $\mathbf{K}_{11}^{(ij)}$, $\mathbf{K}_{12}^{(ij)}$, $\mathbf{K}_{21}^{(ij)}$, $\mathbf{K}_{22}^{(ij)}$ are $6 \times 6$ stiffness matrices. The linear relation (1) should be represented in a global coordinate system.

**Modeling of the rigid link.** If the link flexibility is negligible, the above stiffness model should be replaced by a "rigidity constraint". Applying to this link the rigid body kinematic equation, one can get the following relations between the nodal displacements $\Delta \mathbf{t}_i = [\Delta \mathbf{p}_i; \Delta \boldsymbol{\varphi}_i]$, $\Delta \mathbf{t}_j = [\Delta \mathbf{p}_j; \Delta \boldsymbol{\varphi}_j]$

$$\begin{bmatrix} \mathbf{D}^{(ij)} & -\mathbf{I}_{6 \times 6} \end{bmatrix} \cdot \begin{bmatrix} \Delta \mathbf{t}_i \\ \Delta \mathbf{t}_j \end{bmatrix} = \mathbf{0}_{6 \times 1}; \qquad \mathbf{D}^{(ij)} = \begin{bmatrix} \mathbf{I}_{3 \times 3} & [\mathbf{d}^{(ij)} \times]^T \\ \mathbf{0}_{3 \times 3} & \mathbf{I}_{3 \times 3} \end{bmatrix}_{6 \times 6} \qquad (2)$$

where $[\mathbf{d}^{(ij)} \times]$ denotes the $3 \times 3$ skew-symmetric matrix derived from the vector $\mathbf{d}^{(ij)} = [d_x^{(ij)}, d_y^{(ij)}, d_z^{(ij)}]^T$ describing the link geometry from the $i^{th}$ to $j^{th}$ node.

The second group of equations describing the force equilibrium can be derived using rigid body static equations. Using the above-introduced definition for $\mathbf{D}^{(ij)}$, this equation can be written in a form

$$\begin{bmatrix} \mathbf{I}_{6\times 6} & \mathbf{D}^{(ij)T} \end{bmatrix} \cdot \begin{bmatrix} \mathbf{W}_i \\ \mathbf{W}_j \end{bmatrix} = \mathbf{0}_{6\times 1} \qquad (3)$$

that is also convenient for aggregation of the model components.

## 3. MSA models of manipulator's joints

**Modeling of the rigid joint.** If two adjacent links are connected by means of a rigid joint, the stiffness model must include relations describing two principal rules of structural mechanics: (a) displacement compatibility, (b) force equilibrium. If the nodes corresponding to the adjacent link are denoted as $i, j$, the first of these rules can be expressed as $\Delta\mathbf{t}_i = \Delta\mathbf{t}_j$ yielding the following constraint

$$\begin{bmatrix} \mathbf{I}_{6\times 6} & -\mathbf{I}_{6\times 6} \end{bmatrix}_{6\times 12} \cdot \begin{bmatrix} \Delta\mathbf{t}_i \\ \Delta\mathbf{t}_j \end{bmatrix} = \mathbf{0}_{6\times 1} \qquad (4)$$

The second rule can be written as $\mathbf{W}_i + \mathbf{W}_j = \mathbf{0}$ leading to the second constraint

$$\begin{bmatrix} \mathbf{I}_{6\times 6} & \mathbf{I}_{6\times 6} \end{bmatrix}_{6\times 12} \cdot \begin{bmatrix} \mathbf{W}_i \\ \mathbf{W}_j \end{bmatrix} = \mathbf{0}_{6\times 1} \qquad (5)$$

Similar equations can be written for the rigid connection of three and more links.

**Modeling of the passive joint.** If two adjacent links (*ai*), (*jb*) are connected by means of a passive joint, the displacement compatibility condition must be replaced by $\mathbf{\Lambda}_{ij}^r \cdot (\Delta\mathbf{t}_i - \Delta\mathbf{t}_j) = \mathbf{0}$, where $\mathbf{\Lambda}_{ij}^r$ is a rank-deficient matrix that defines directions for the passive joint that do not admit free relative motions. This allows presenting the first constraint in the form

$$\begin{bmatrix} \mathbf{\Lambda}_{*ij}^r & -\mathbf{\Lambda}_{*ij}^r \end{bmatrix}_{r\times 12} \cdot \begin{bmatrix} \Delta\mathbf{t}_i \\ \Delta\mathbf{t}_j \end{bmatrix} = \mathbf{0}_{r\times 1} \qquad (6)$$

that contains a rectangular matrix $\mathbf{\Lambda}_{*ij}^r$ of the rank $r$. In the simple cases, the matrix $\mathbf{\Lambda}_{*ij}^r$ can be easily derived from $\mathbf{\Lambda}_{ij}^r$ by simple elimination of zero rows.

The force equilibrium condition is presented in this case as $\mathbf{\Lambda}_{ij}^r \cdot (\mathbf{W}_i + \mathbf{W}_j) = \mathbf{0}$ that gives the second constraint in the following form

$$\begin{bmatrix} \mathbf{\Lambda}_{*ij}^r & \mathbf{\Lambda}_{*ij}^r \end{bmatrix}_{r\times 12} \cdot \begin{bmatrix} \mathbf{W}_i \\ \mathbf{W}_j \end{bmatrix} = \mathbf{0}_{r\times 1} \qquad (7)$$

In addition, passive joints do not transmit the force/torque in the direction corresponding to zero columns in the matrix $\mathbf{\Lambda}_{ij}^r$, which yields complementary equations

$$\mathbf{\Lambda}_{*ij}^p \cdot \mathbf{W}_i = \mathbf{0}_{p\times 1}; \qquad \mathbf{\Lambda}_{*ij}^p \cdot \mathbf{W}_j = \mathbf{0}_{p\times 1} \qquad (8)$$

where $\mathbf{\Lambda}_{*ij}^p$ denotes the full-rank matrix of size $p\times 6$ defining passive directions, and $p = 6 - r$.

**Modeling of the elastic joint.** If two adjacent links are connected by an elastic joint, that can be treated as a passive joint with springs. In this case, the deflection compatibility condition remains the same. However, the force equilibrium must be slightly revised and presented as

$$\mathbf{W}_i + \mathbf{W}_j = \mathbf{0}; \qquad \mathbf{\Lambda}_{*ij}^e \cdot \mathbf{W}_i = \mathbf{K}_{ij}^e \cdot \mathbf{\Lambda}_{*ij}^e \cdot (\Delta \mathbf{t}_i - \Delta \mathbf{t}_j) \qquad (9)$$

where the matrix $\mathbf{\Lambda}_{*ij}^e$ of size $e\times 6$, $e = 6 - r$ corresponds to the non-rigid directions of the joint (similar to $\mathbf{\Lambda}_{*ij}^p$) and $\mathbf{K}_{ij}^e$ is $e\times e$ stiffness matrix describing elastic properties of the joint. In the compact form, the letter equations can be presented as

$$\begin{bmatrix} \mathbf{0}_{6\times 6} & \mathbf{0}_{6\times 6} & \mathbf{I}_{6\times 6} & \mathbf{I}_{6\times 6} \\ -\mathbf{K}_{ij}^e \cdot \mathbf{\Lambda}_{*ij}^e & \mathbf{K}_{ij}^e \cdot \mathbf{\Lambda}_{*ij}^e & \mathbf{\Lambda}_{*ij}^e & \mathbf{0}_{e\times 6} \end{bmatrix}_{(6+e)\times 24} \cdot \begin{bmatrix} \Delta \mathbf{t}_i \\ \Delta \mathbf{t}_j \\ \mathbf{W}_i \\ \mathbf{W}_i \end{bmatrix}_{24\times 1} = \mathbf{0}_{(6+e)\times 1} \qquad (10)$$

It should be stressed that the above expressions are valid for the so-called "no preloading case", when equal deflections do not generate the elastic forces, i.e. $\mathbf{\Lambda}_{*ij}^e \cdot \mathbf{W}_i = \mathbf{\Lambda}_{*ij}^e \cdot \mathbf{W}_j = \mathbf{0}$. In the case when the springs are initially preloaded by the wrench $\mathbf{W}_{ij}^0$, the force equilibrium equation must be presented in the form

$$\mathbf{W}_i + \mathbf{W}_j = \mathbf{0}; \qquad \mathbf{\Lambda}_{*ij}^e \cdot (\mathbf{W}_i - \mathbf{W}_{ij}^0) = \mathbf{K}_{ij}^e \cdot \mathbf{\Lambda}_{*ij}^e \cdot (\Delta \mathbf{t}_i - \Delta \mathbf{t}_j) \qquad (11)$$

and the right-hand side of (10) should be slightly modified.

**Modeling of the actuated joint.** The actuated joint ensures transmission of the force/torque between the manipulator links. In the frame of MSA technique, it can be presented either as a rigid, passive or elastic connection. In the first two cases, the actuating effort is transmitting in the direction corresponding to the vectors $\mathbf{u}_1, ..., \mathbf{u}_r$ while in the last case transmission can be performed in the direction defined by the vectors $\mathbf{u}_{r+1}, ..., \mathbf{u}_6$. Hence, the actuated joint may be included in the global stiffness model as a set of two linear matrix constraints.

## 4. Including of boundary conditions and loadings

**Modeling of rigid support.** The rigid connection of the link (*jb*) to the robot base can be presented as a special case of the rigid joint with an eliminated link (*ai*) and zero deflection $\Delta \mathbf{t}_j = \mathbf{0}$. This simplifies the deflection compatibility constraint as

$$\left[\mathbf{I}_{6\times 6}\right] \cdot \left[\Delta \mathbf{t}_j\right] = \mathbf{0}_{6\times 1} \qquad (12)$$

that contains 6 linear equations to be included in the global stiffness model. Corresponding reaction wrench at the rigid support may be computed as

$$\mathbf{W}_j = \mathbf{K}_{12}^{(jb)} \cdot \Delta \mathbf{t}_b \qquad (13)$$

where the deflection $\Delta \mathbf{t}_b$ is obtained from the solution of the global stiffness equations.

**Modeling of passive support.** The passive connection of the link (*jb*) to the robot base can be presented as a special case of the passive joint with an eliminated link (*ai*) and zero deflection in the non-passive directions $\mathbf{\Lambda}_{*ij}^r \cdot \Delta \mathbf{t}_j = \mathbf{0}$. This simplifies the deflection compatibility constraint down to

$$\left[\mathbf{\Lambda}_{*ij}^r\right]_{r\times 6} \cdot \left[\Delta \mathbf{t}_j\right] = \mathbf{0}_{r\times 1} \qquad (14)$$

that contributes $r$ linear equations to the global stiffness model. In addition, it is necessary to take into account that the wrench components for the passive direction are equal to zero

$$\mathbf{\Lambda}_{*ij}^p \cdot \mathbf{W}_j = \mathbf{0} \qquad (15)$$

that contributes another $p$ equations to the global stiffness model. Totally, this gives $r + p = 6$ independent linear constraints.

**Modeling of elastic support.** Similarly, the elastic connection of the link (*jb*) to the robot base can be presented as a special case of the elastic joint with an eliminated link (*ai*) and zero deflection in the non-elastic directions $\mathbf{\Lambda}_{*ij}^r \cdot \Delta \mathbf{t}_j = \mathbf{0}$. This allows us to use the same deflection compatibility constraint as above, which contributes $r$ linear equations to the global stiffness model. In addition, it is necessary to consider that the wrench components corresponding to the non-rigid directions are produced by the elastic forces satisfying the Hook's law $\mathbf{\Lambda}_{*ij}^e \cdot \mathbf{W}_j = \mathbf{K}_{ij}^e \cdot \mathbf{\Lambda}_{*ij}^e \cdot \Delta \mathbf{t}_j$. The latter leads to the linear relation

$$\left[\mathbf{K}_{ij}^e \cdot \mathbf{\Lambda}_{*ij}^e \;\middle|\; -\mathbf{\Lambda}_{*ij}^e\right]_{e\times 12} \cdot \begin{bmatrix}\Delta \mathbf{t}_j \\ \mathbf{W}_j\end{bmatrix} = \mathbf{0}_{e\times 1} \qquad (16)$$

that contributes another $e$ scalar equations to the global stiffness model. Totally, this gives $r + e = 6$ independent linear constraints.

**Including the external loading**. In robotic manipulators, both serial and parallel, there is at least one node that is not connected directly to the robot base. It corresponds to the end-effector that interacts with robot environment by applying the force/torque to the external objects. For the global stiffness model, the end-effector produces the boundary conditions that should include the vector of the external wrench $\mathbf{W}_e$, which is also necessary for computation of the Cartesian stiffness matrix.

To take into account the external loading, the global stiffness model must be completed by the linear constraint derived from the force equilibrium at the node $e$, i.e. $\mathbf{W}_i + \mathbf{W}_j = \mathbf{W}_e$, which can be rewritten in the form

$$\begin{bmatrix} \mathbf{I}_{6\times 6} & \mathbf{I}_{6\times 6} \end{bmatrix}_{6\times 12} \cdot \begin{bmatrix} \mathbf{W}_i \\ \mathbf{W}_j \end{bmatrix}_{12\times 1} = \begin{bmatrix} \mathbf{W}_e \end{bmatrix}_{6\times 1} \qquad (17)$$

Similarly, it is possible to derive the constraints produced by the external wrenches applied to other nodes.

## 5. Aggregation of MSA model components

In contrast to large mechanical structures consisting of a huge number of flexible components, the robotic manipulator is rather simple for MSA analysis. Since the number of flexible links in manipulator is relatively small, the assembling stage can be simplified and the columns/rows merging operations can be avoided and replaced by adding relevant constraints.

As it was shown above, the MSA equations for robotic manipulator are derived from three main sources: (i) link models, (ii) joint models and (iii) boundary conditions. The first of them describes the force-displacement relations for all links (both flexible and rigid) yielding 12 scalar equations per link relating 24 variables $\{\Delta \mathbf{t}_i, \Delta \mathbf{t}_j, \mathbf{W}_i, \mathbf{W}_j\}$. The second group of equations ensures the displacement compatibility and force/torque equilibrium for each internal connection (both rigid, passive and elastic), it includes two types of relations written for $\{\Delta \mathbf{t}_i, \Delta \mathbf{t}_j,...\}$ and $\{\mathbf{W}_i, \mathbf{W}_j,...\}$ separately. Independent of the connection type, each joint provides 12 scalar equations if it connects two links, 18 scalar equations for the connection of three links, etc. The third group of equations is issued from the manipulator connections to the environment, they are presented as the force/displacement constraints for certain nodes that give up to 6 scalar equations for $\{\Delta \mathbf{t}_i, \Delta \mathbf{t}_j,...\}$ or at least 6 scalar equations for $\{\mathbf{W}_i, \mathbf{W}_j,...\}$ depending on the connection type.

To derive the global stiffness model of the considered manipulator, let us combine all displacements and all wrenches in single vectors and denote them as $\{\Delta \mathbf{t}_i\}$ and $\{\mathbf{W}_i\}$. This allows us to present the desired model in the form of the following block-matrix equation of sparse structure

$$\begin{array}{l}
\forall \;\; flexible \;\; links \\
\forall \;\; rigid \;\; links \\
\\
\forall \;\; rigid \;\; joints \\
\\
\forall \;\; passive \;\; joints \\
\\
\forall \;\; elastic \;\; joints \\
\\
\forall \;\; rigid \;\; supports \\
\forall \;\; passive \;\; supports \\
\\
\forall \;\; elastic \;\; supports \\
\\
external \;\; loading
\end{array}
\begin{bmatrix}
\{\mathbf{K}_{12\times12}^{ij}\} & \{-\mathbf{I}_{12\times12}\} \\
\{\mathbf{D}_{6\times6}^{ij},-\mathbf{I}_{6\times6}\} & \\
 & \{\mathbf{I}_{6\times6},\mathbf{D}_{6\times6}^{ij\,T}\} \\
\{\mathbf{I}_{6\times6},-\mathbf{I}_{6\times6}\} & \\
 & \{\mathbf{I}_{6\times6},\mathbf{I}_{6\times6}\} \\
\{\Lambda_{*ij}^r,-\Lambda_{*ij}^r\} & \\
 & \{\Lambda_{*ij}^r,\Lambda_{*ij}^r\} \\
 & \{\Lambda_{*ij}^p\} \\
\{\Lambda_{*ij}^r,-\Lambda_{*ij}^r\} & \\
 & \{\mathbf{I}_{6\times6},\mathbf{I}_{6\times6}\} \\
\{\mathbf{K}_{ij}^e\Lambda_{*ij}^e,-\mathbf{K}_{ij}^e\Lambda_{*ij}^e\} & \Lambda_{*ij}^e \\
\{\mathbf{I}_{6\times6}\} & \\
\{\Lambda_{*ij}^r\} & \\
 & \{\Lambda_{*ij}^p\} \\
\{\Lambda_{*ij}^r\} & \\
\{\mathbf{K}_{ij}^e\Lambda_{*ij}^e\} & \Lambda_{*ij}^e \\
 & \{-\mathbf{I}_{6\times6}\}
\end{bmatrix}
\begin{bmatrix}\{\Delta\mathbf{t}_i\}\\ \{\mathbf{W}_i\}\end{bmatrix}=
\begin{bmatrix}
0\\0\\0\\0\\\{\mathbf{W}_i^{ext}\}\\0\\\{\mathbf{W}_{*i}^{ext}\}\\0\\0\\\{\mathbf{W}_i^{ext}\}\\\{\mathbf{W}_{ij}^0\}\\0\\0\\0\\0\\\{\mathbf{W}_{ij}^0\}\\\{\mathbf{W}_i^{ext}\}
\end{bmatrix} \quad (18)$$

After rearranging the matrix rows and introducing relevant definitions for the blocks, the global stiffness model can be presented as

$$\begin{bmatrix}\mathbf{K}_{agr} & \mathbf{S}_{agr}\\ \mathbf{A}_{agr} & 0\\ 0 & \mathbf{B}_{agr}\end{bmatrix}\cdot\begin{bmatrix}\Delta\mathbf{t}_{agr}\\ \mathbf{W}_{agr}\end{bmatrix}=\begin{bmatrix}\mathbf{W}_0\\ 0\\ \mathbf{W}_{ext}\end{bmatrix} \quad (19)$$

Further, separating the node variables $\{\Delta\mathbf{t}_i\}$ in two groups corresponding internal nodes $\Delta\mathbf{t}_m$ and to the end effector $\Delta\mathbf{t}_e$ one can get the following linear system

$$\begin{bmatrix}\mathbf{S}_{agr} & \mathbf{K}_{agr}^m & \mathbf{K}_e\\ 0 & \mathbf{A}_{agr}^m & \mathbf{A}_e\\ \mathbf{B}_{agr}^m & 0 & 0\\ \mathbf{B}_e & 0 & 0\end{bmatrix}\cdot\begin{bmatrix}\mathbf{W}_{agr}\\ \Delta\mathbf{t}_m\\ \Delta\mathbf{t}_e\end{bmatrix}=\begin{bmatrix}\mathbf{W}_0\\ 0\\ \mathbf{W}_{ext}^m\\ \mathbf{W}_e\end{bmatrix} \quad (20)$$

that allows expressing the external wrench $\mathbf{W}_e$ with respect to $\Delta\mathbf{t}_e$ as

$$\mathbf{W}_e=-\begin{bmatrix}\mathbf{B}_e & 0\end{bmatrix}\begin{bmatrix}\mathbf{S}_{agr} & \mathbf{K}_{agr}^m\\ 0 & \mathbf{A}_{agr}^m\\ \mathbf{B}_{agr}^m & 0\end{bmatrix}^{-1}\begin{bmatrix}\mathbf{K}_e\\ \mathbf{A}_e\\ 0\end{bmatrix}\cdot\Delta\mathbf{t}_e+\begin{bmatrix}\mathbf{B}_e & 0\end{bmatrix}\begin{bmatrix}\mathbf{S}_{agr} & \mathbf{K}_{agr}^m\\ 0 & \mathbf{A}_{agr}^m\\ \mathbf{B}_{agr}^m & 0\end{bmatrix}^{-1}\begin{bmatrix}\mathbf{W}_0\\ 0\\ \mathbf{W}_{ext}^m\end{bmatrix} \quad (21)$$

and to obtain the desired Cartesian stiffness matrix.

## 6. Conclusion

The paper deals with the advancement of the Matrix structural analysis for robotics applications. In contrast to previous results, it is suitable for robots with flexible links, rigid connections, passive and elastic joints with external loadings and preloadings. The paper describes in detail required matrix transformations that allow the user to obtain desired force-deflection relation and torque/deflections in internal mechanical elements. Its allows user straightforward aggregation of stiffness model equations avoiding traditional column/row merging procedures in the extended stiffness matrix.

**Acknowledgments.** The work presented in this paper was supported by the grant of Russian Science Foundation №17-19-01740.